\documentclass[final,5p,times]{elsarticle}

\usepackage{amssymb}
\usepackage{amsmath}

\usepackage{amsmath}
\usepackage{xspace}
\usepackage{todonotes}

\usepackage{siunitx}

\usepackage[english]{babel}

\usepackage[utf8]{inputenc}

\usepackage{caption}
\usepackage{tabularx} 

\usepackage{booktabs}

\graphicspath{{./images/}}

\usepackage{enumitem}

\usepackage{graphicx}
\usepackage{subcaption}

\usepackage[singlespacing]{setspace}

\usepackage{natbib}
\setcitestyle{numbers}

\usepackage{rotating}
\usepackage{pdflscape,kantlipsum}

\usepackage{makecell}
\usepackage{ifoddpage}

\usepackage{xcolor}

\usepackage{amsfonts}

\usepackage[bookmarks,bookmarksnumbered]{hyperref}
\hypersetup{colorlinks = true,linkcolor = blue,anchorcolor =red,citecolor = blue,filecolor = red,urlcolor = red,
            pdfauthor=author}
\usepackage{xurl}
\usepackage{microtype}

\usepackage{amssymb}
\usepackage{amsthm}
\usepackage{multirow}

\journal{...}

\begin{document}

\begin{frontmatter}



\title{DefTransNet: A Transformer-based Method for Non-Rigid Point Cloud Registration in the Simulation of Soft Tissue Deformation}

\author[1]{Sara Monji-Azad\corref{cor1}} 
\author[2]{Marvin Kinz}
\author[8]{Siddharth Kothari}
\author[1]{Robin Khanna}
\author[1]{Amrei Carla Mihan}
\author[3]{David M{\"a}nnel} 
\author[3,4]{Claudia Scherl}     
\author[1,5,6,7]{J\"urgen Hesser}     
    
\cortext[cor1]{Corresponding author: Tel.: +496213838186; Email address: sara.monjiazad@medma.uni-heidelberg.de\\}

\address[1]{Mannheim Institute for Intelligent Systems in Medicine (MIISM), Medical Faculty Mannheim, Heidelberg University, Mannheim, Germany}

\address[2]{Department of Radiation Oncology, Brigham and Women's Hospital, Dana-Farber Cancer Institute, Harvard Medical School, Boston, MA, USA}

\address[8]{International Institute of Information Technology, Bangalore, India}

\address[3]{Department of Otorhinolaryngology, Head and Neck Surgery, Medical Faculty Mannheim, Heidelberg University, Mannheim, Germany}

\address[4]{AI Health Innovation Cluster, Heidelberg-Mannheim Health and Life Science Alliance, Heidelberg, Germany}

\address[5]{Interdisciplinary Center for Scientific Computing (IWR), Heidelberg University, Heidelberg, Germany}

\address[6]{Central Institute for Computer Engineering (ZITI), Heidelberg University, Heidelberg, Germany}

\address[7]{CZS Heidelberg Center for Model-Based AI, Heidelberg University, Mannheim, Germany}


\begin{abstract}
Soft-tissue surgeries, such as tumor resections, are complicated by tissue deformations that can obscure the accurate location and shape of tissues. By representing tissue surfaces as point clouds and applying non-rigid point cloud registration (PCR) methods, surgeons can better understand tissue deformations before, during, and after surgery. Existing non-rigid PCR methods, such as feature-based approaches, struggle with robustness against challenges like noise, outliers, partial data, and large deformations, making accurate point correspondence difficult. Although learning-based PCR methods, particularly Transformer-based approaches, have recently shown promise due to their attention mechanisms for capturing interactions, their robustness remains limited in challenging scenarios. In this paper, we present DefTransNet, a novel end-to-end Transformer-based architecture for non-rigid PCR. DefTransNet is designed to address the key challenges of deformable registration—including large deformations, outliers, noise, and partial data—by inputting source and target point clouds and outputting displacement vector fields. The proposed method incorporates a learnable transformation matrix to enhance robustness to affine transformations, integrates global and local geometric information, and captures long-range dependencies among points using Transformers. We validate our approach on four datasets: ModelNet, SynBench, 4DMatch, and DeformedTissue, using both synthetic and real-world data to demonstrate the generalization of our proposed method. Experimental results demonstrate that DefTransNet outperforms current state-of-the-art registration networks across various challenging conditions. Our code and data are publicly available at \url{https://https://github.com/m-kinz/DefTransNet} and \url{https://doi.org/10.11588/data/R9IKCF              }.
\end{abstract}








\end{frontmatter}


\renewcommand{\arraystretch}{1.5} 

\section{Introduction} \label{introduction}
Tissue deformation is a critical issue in soft-tissue surgery. It occurs after wound opening due to factors like tension loss, patient positioning, or tissue extraction. This effect has been particularly noted in head and neck surgery, especially during tumor resection, where critical structures are closely located and may require protection \cite{monji2024point,mannle2023artificial}.

Point cloud registration is a key aspect of computer vision, focused on estimating transformations to align sets of corresponding points \cite{monji2023review}. This technique is essential in a variety of fields, including virtual and augmented reality \cite{mahmood20193d}, LiDAR-based applications \cite{wang2018lidar}, and quality control in manufacturing \cite{wang2019applications}. The aim of point cloud registration is to reduce alignment errors between transformed and target point clouds \cite{deng2022survey}. Registration methods can generally be divided into two types: rigid and non-rigid transformations. Rigid registration applies affine transformations like rotation and translation, preserving the object’s geometry and shape, while non-rigid registration determines a deformation field to match the source and target point clouds \cite{monji2023review}. The central challenges include identifying stable corresponding points and devising an accurate transformation function, while also ensuring robustness against variations in deformation, noise, outliers, and incomplete data \cite{Castellani.2020, wang2022gp}. Recent advancements in non-rigid point cloud registration often categorize methods as either coarse or fine, with another classification separating feature-based methods—focused on matching distinctive features—from those that directly estimate the deformation field \cite{deng2022survey}.

Both rigid and non-rigid registration approaches can further be divided into non-learning and learning-based methods. Non-learning techniques rely primarily on iterative optimization to derive the transformation, using predefined mathematical models and metrics to align point clouds \cite{Horn.2020}. For instance, algorithms such as Iterative Closest Point (ICP) and its variants iteratively minimize the distance between corresponding points, achieving reliable registration in many cases \cite{besl1992method, jiang2021review}. In contrast, learning-based techniques employ machine learning, particularly deep learning, to infer transformations directly from point cloud features. These methods utilize neural networks to capture complex point relationships, accommodating a wider range of deformations and noise levels \cite{huang2021comprehensive}. Typically, learning-based approaches are trained on large datasets, allowing them to learn feature representations and correspondences that generalize well to unseen data \cite{bauer2021reagent}. The primary differences between non-learning and learning-based methods lie in their modeling techniques, adaptability to complex deformations, resilience to noise and outliers, computational requirements, and reliance on training data \cite{sun2022weakly}. Non-learning methods can be applied directly without training data, while learning-based methods require extensive labeled datasets for training and often involve higher computational demands \cite{ao2021spinnet}.

In recent years among all learning-based PCR techniques, Transformers have become more popular \cite{monji2023review}. Transformers, introduced by Vaswani et al. \cite{vaswani2017attention}, designed for sequential data, have gained popularity in domains such as computer vision and natural language processing \cite{devlin2018bert,dosovitskiy2020image}. While widely used for classification and segmentation tasks \cite{zhao2021point,guo2021pct}, their application in PCR is limited. Early methods like Deep Closest Point \cite{wang2019deep} employ Transformers for rigid PCR using embedding, attention mechanisms, and differentiable singular value decomposition. Some newer approaches introduce the Transformer architecture into the network \cite{qin2022geometric,yu2021cofinet,yang2022one}, commonly using self-attention and cross-attention interchangeably \cite{yew2022regtr}. These methods leverage interactions between features to enhance feature representation. In Lepard \cite{li2022lepard}, explicit position encoding for learning 3D relative distance information is introduced. In Regtr \cite{yew2022regtr}, an end-to-end framework with stacked cross-encoder layers is presented. Yu et al. \cite{yu2021cofinet} propose CoFiNet, a coarse-to-fine pipeline with a deeper Transformer network at the coarse level. Subsequently, Qin et al. \cite{qin2022geometric} introduce an effective position encoding method in the self-attention module to capture geometric information. \par

However, Transformer-based registration methods still face a challenge known as feature ambiguity \cite{min2021distinctiveness}, where the network mistakenly matches similar features from unrelated points. Feature ambiguity can be expressed mathematically as the uncertainty in establishing correspondences between points in different point clouds. Let us denote a point $x_i$ in source point cloud $X$ and its corresponding point $y_j$ in target point cloud $Y$ under a non-rigid transformation $T$, then the feature ambiguity $A(x_i)$ is given by the minimum matching cost or dissimilarity measure $A(x_i) = \min_{y_j} C(x_i, y_j)$. Here, $C(x_i, y_j)$ represents the cost or dissimilarity measure between the corresponding points. Attempts to address feature ambiguity involve modifying the self-attention part, including elements like oriented positional equilibrium \cite{min2021distinctiveness}, position encoding \cite{li2022lepard}, and geometric information \cite{qin2022geometric}. Unfortunately, these solutions either introduce significant computational overhead or lack robustness in challenging scenarios \cite{min2021distinctiveness}. 

To address and solve the mentioned problem, DefTransNet is introduced in this paper. DefTransNet is a novel end-to-end network architecture based on Transformers, where source and target point clouds are the input, and displacement vector fields are the output. The method involves two main steps: first, a unique feature descriptor based on the Transformers, and second, learning the displacement vector field. The feature descriptor includes a learnable transformation matrix to be robust to affine transformations, integrates global and local geometric information for feature aggregation, and incorporates a Transformer to capture global context and relationships between points. The main contributions of our work are summarized as follows:

\begin{itemize}
    \item We introduce DefTransNet, a Transformer-based feature descriptor designed to improve feature disambiguation in non-rigid point cloud registration.
    \item Our model uses identical encoder and decoder blocks, incorporating multi-head self-attention and feedforward networks to capture global context and identify relationships between points, making it robust to deformations, noise, outliers, partial data, and rotations.
    \item We demonstrate the effectiveness of Transformers in capturing information from both source and target features using a sequence-to-sequence approach.
    \item We validate DefTransNet on four synthetic and real-world datasets to demonstrate the generalization of our method.
\end{itemize}

The organization of this paper is as follows: Section \ref{RelatedWork} reviews related work on non-rigid point cloud registration, and Section \ref{ProblemDefinition} introduces the problem definition. Our proposed method is described in Section \ref{ProposedMethod}. Section \ref{Evaluation} evaluates the method on various datasets, highlighting their characteristics and comparing our approach to baseline methods. In Section \ref{Discussion}, we discuss key aspects of our method across different datasets, and the paper concludes in Section \ref{Conclusion}.

\section{Related Work}\label{RelatedWork}
In recent years, the use of transformers and DGCNN-based approaches for non-rigid point cloud registration has increased. This section presents a review of several registration methods based on these two approaches. A summary of some notable methods is provided in Table \ref{table:LearningMethod}.

\begin{table*}
\footnotesize
\centering
\caption{Overview of some learning-based non-rigid point cloud registration methods based on Transformers and DGCNN}
\label{table:LearningMethod}
\begin{tabular}{p{2.25cm} p{1cm} p{2.5cm} p{4cm} p{6cm}} 
\hline
Methods&Year& \raggedright Network Architecture & Robustness & Experimental Data \\
\hline

\citep{hansen2019learning}&2019& DGCNN & Noise, outliers & Medical dataset \\
\hline

PRNet \citep{wang2019prnet}&2019 & \raggedright DGCNN, Transformer  & Noise, partial & ShapeNetCore\citep{chang2015shapenet}, ModelNet40 \citep{wu20153d}  \\
\hline

CoFiNet \cite{yu2021cofinet}&2021& Transformer& Outlier, partial & odometryKITTI \cite{geiger2013vision}, 3Dmatch \cite{zeng20173dmatch}, 3DLoMatch \cite{huang2021predator}\\\hline

NrtNet \citep{hu2022nrtnet}&2022  &  \raggedright DGCNN, Transformer & Deformation levels & SURREAL \citep{varol2017learning}, SHREC’19 \citep{melzi2019shrec}, MIT \citep{grosse2009ground}\\
\hline

\cite{qin2022geometric}&2022&Transformer&Outlier, partial&odometryKITTI \cite{geiger2013vision}, 3Dmatch \cite{zeng20173dmatch}, 3DLoMatch \cite{huang2021predator}\\\hline

\cite{yew2022regtr} &2022&Transformer&Outlier, partial&3Dmatch \cite{zeng20173dmatch}, 3DLoMatch \cite{huang2021predator}, ModelNet \cite{wu20153d}\\\hline
Lepard \cite{li2022lepard} &2022&Transformer&Outlier, partial, deformation levels&3Dmatch \cite{zeng20173dmatch}, 3DLoMatch \cite{huang2021predator}, 4DMatch \cite{li2022lepard} \\\hline

OIF-PCR \cite{yang2022one} &2022&Transformer&Outlier, partial&odometryKITTI \cite{geiger2013vision}, 3Dmatch \cite{zeng20173dmatch}, 3DLoMatch \cite{huang2021predator}\\\hline

GraphSCNet \cite{qin2023deep}&2023&GCNN&Outliers, deformation levels, partial & 4DMatch \cite{li2022lepard}, CAPE \cite{pons2017clothcap}, DeepDeform \cite{bozic2020deepdeform}\\
\hline

NIE \cite{jiang2023neural}&2023&DGCNN&Noise, partial &SURREAL \citep{varol2017learning}, FAUST \cite{bogo2014faust}, SCAPE \cite{anguelov2005scape}\\
\hline

MAFNet \cite{chen2024mafnet}&2024&Transformer&Noise, partial&7-Scenes \citep{shotton2013scene}, ModelNet \citep{wu20153d}\\
\hline

Robust-DefReg\cite{monji2024robust} &2024&GCNN&Noise, outliers, deformation levels&SynBench\cite{DataSynBench}, ModelNet \citep{wu20153d}\\
\hline

\end{tabular}
\end{table*}

Transduction models refer to machine learning models that convert one type of data (usually a sequence) into another type, while preserving the input-output structure \cite{belhasin2022transboost}. In the context of sequence transduction, the goal is to map an input sequence (e.g., a sentence in one language) to an output sequence (e.g., the translated sentence in another language). Transduction models typically use architectures like Recurrent Neural Networks (RNNs), Convolutional Neural Networks (CNNs), or Transformers, which are designed to handle sequential data efficiently. Most sequence transduction models use complex recurrent or convolutional neural networks with an encoder-decoder structure, often improved by attention mechanisms \cite{chalvidal2024learning}. This setup is particularly effective for tasks such as machine translation, where maintaining contextual relevance between input and output is crucial.

In \cite{vaswani2017attention}, the transformer is introduced as a model in the field of deep learning, particularly for natural language processing (NLP) tasks like machine translation. The authors propose the transformer, an architecture that relies solely on attention mechanisms, eliminating the need for recurrence or convolutions entirely. This innovation allows for faster training through parallelization and easier modeling of long-range dependencies in sequences. This paper became the foundation for subsequent advancements, including BERT, GPT, and other transformer-based models that dominate NLP today. Notably, the flexibility and power of transformers have led to their adaptation beyond language tasks, with applications in 3D point cloud processing. For instance, among the initial studies using transformers to process point clouds, \cite{zhao2021point} introduces an approach to 3D point cloud processing by adapting self-attention mechanisms, which have been highly successful in natural language processing and image analysis. The authors design self-attention layers specifically for 3D point clouds and construct networks to tackle tasks like semantic scene segmentation, object part segmentation, and object classification. Further expanding on this, \cite{guo2021pct} presents a Transformer-based framework for processing 3D point clouds, addressing challenges like their irregular domain and lack of order. Unlike traditional deep learning methods, PCT is inherently permutation invariant, making it well-suited for unordered point cloud data. To capture local geometric context effectively, the framework enhances input embedding through farthest point sampling and nearest neighbor search. PCT demonstrates state-of-the-art performance across various tasks, including shape classification, part segmentation, semantic segmentation, and normal estimation, showcasing its effectiveness and versatility for point cloud learning in applications such as autonomous driving, robotics, and 3D modeling.

While transformers have seen widespread use in classification and segmentation tasks, their application in point cloud registration (PCR) remains limited. Early methods like Deep Closest Point \cite{wang2019deep} employ transformers for rigid PCR, incorporating three main components: a point cloud embedding network to extract meaningful features, an attention-based module with a pointer generation layer for estimating combinatorial matches between points, and a differentiable singular value decomposition (SVD) layer to compute the rigid transformation. To address the challenges of point cloud registration, \cite{qin2022geometric} introduces a Geometric Transformer that improves the extraction of accurate correspondences in low-overlap scenarios, bypassing traditional keypoint detection. This keypoint-free approach matches superpoints (downsampled clusters of points) based on the overlap of their neighboring patches, further enhancing robustness and ensuring invariance to rigid transformations. Similarly, CoFiNet (Coarse-to-Fine Network) \cite{yu2021cofinet} tackles point cloud registration without relying on keypoint detection. It operates in a hierarchical manner, first matching down-sampled nodes at a coarse scale, where the model uses a weighting scheme to focus on areas with more overlap, thus reducing the search space for the next stage. At a finer scale, the model refines these matches by expanding node proposals into patches of points with associated descriptors. A density-adaptive matching module is then used to further refine point correspondences based on overlap in corresponding patches, effectively handling varying point densities. These techniques highlight the ongoing efforts to enhance the accuracy and efficiency of point cloud registration, particularly when facing complex real-world challenges such as varying densities and overlapping regions.

Another notable contribution in point cloud registration is OIF-PCR \cite{yang2022one}, a position encoding method that improves correspondence accuracy by using one inlier as a reference. The method finds a single correspondence through a differentiable optimal transport layer, then normalizes each point for position encoding. This approach addresses issues related to differing reference frames between point clouds and reduces feature ambiguity by learning spatial consistency. The integration of correspondence and position encoding in an iterative optimization process allows for progressive refinement of point cloud alignment and feature learning. Along these lines, \cite{yew2022regtr} proposes an end-to-end point cloud registration framework that eliminates the need for traditional feature matching and outlier filtering methods like RANSAC, replacing them with attention mechanisms. The network, built around transformer layers, incorporates self- and cross-attention mechanisms to directly predict the final set of correspondences, which can then be used to estimate the required rigid transformation without the need for post-processing steps.

For partial point cloud matching, Lepard \cite{li2022lepard} integrates 3D positional knowledge into the registration process. This method begins by combining the KPFCN feature extractor with the Transformer model and differentiable matching techniques, and introduces three key innovations to enhance the use of 3D positional information. These include disentangling point cloud representations into feature and position spaces, developing a position encoding method to capture 3D relative distances, and implementing a repositioning module that adjusts relative positions between points across point clouds. These improvements enable more accurate matching in partial point cloud registration tasks, further illustrating the potential of transformer-based methods for dealing with complex point cloud data.

Shifting focus to Graph Convolutional Neural Networks (GCNNs), \cite{zhang2019graph} shows how these networks process data structured as graphs, learning geometric features based on the relationships between neighboring nodes. This approach is particularly effective for point clouds, where the structure of data is inherently graph-like. In the medical domain, GCNNs have shown promise for tasks like 3D lung registration \cite{hansen2021deep}, where edge convolutions are used to extract geometric features, and Loopy Belief Propagation (LBP) regularizes displacements on a k-nearest neighbor graph. Additionally, \cite{hansen2019learning} introduces a dynamic GCNN approach for point cloud registration, which refines correspondences probabilistically using the Coherent Point Drift (CPD) algorithm. These approaches demonstrate the versatility of GCNNs in handling various point cloud registration challenges, particularly when dealing with complex deformable structures.

Continuing this line of research, \cite{hu2022nrtnet} introduces NrtNet, an unsupervised transformer-based network designed for non-rigid point cloud registration. This method leverages self-attention mechanisms to extract feature correspondences between large deformations. NrtNet's three main components—a feature extraction module, a correspondence matrix generation module, and a reconstruction module—work together to align point clouds by learning and normalizing correspondence probabilities. This approach is designed to handle large-scale deformations, a significant challenge in non-rigid point cloud registration. Extending this work, \cite{qin2023deep} proposes GraphSCNet, a network that tackles outlier correspondence pruning, particularly in non-rigid point cloud registration. GraphSCNet addresses the challenge of local rigidity in non-rigid deformations by using a local spatial consistency measure to evaluate correspondence compatibility, ensuring better outlier discrimination and improving the overall accuracy of registration.

Building on the concept of structural alignment, \cite{jiang2023neural} presents NIE, a method for embedding vertices of point clouds into a high-dimensional space to preserve intrinsic structural properties. This technique is particularly useful for aligning point clouds sampled from deformable shapes, which often lack explicit structural information. NIE forms the foundation for a weakly-supervised framework for non-rigid point cloud registration, avoiding expensive preprocessing steps and the reliance on ground-truth correspondence labels. Finally, Robust-DefReg \cite{monji2024robust} leverages graph convolutional networks in a coarse-to-fine framework to handle large deformations, noise, and outliers. This end-to-end approach focuses on global feature learning to establish accurate correspondences and transformations across varying deformation scales. Another key contribution of this work is the development of SynBench \cite{DataSynBench}, a comprehensive benchmark dataset for evaluating non-rigid registration methods, which supports further advancements in the field.

\section{Problem Definition of Non-Rigid Point Cloud Registration} \label{ProblemDefinition}

Let \( \mathbf{X} = \{ \mathbf{x}_i \in \mathbb{R}^3 \mid i = 1, \ldots, N \} \) and \( \mathbf{Y} = \{ \mathbf{y}_j \in \mathbb{R}^3 \mid j = 1, \ldots, M \} \) be two sets of 3D points. The goal of non-rigid point cloud registration is to find a transformation \( \mathcal{T} \) that deforms the source point cloud \( \mathbf{X} \) to align it with the target point cloud \( \mathbf{Y} \). This transformation maps each point \( \mathbf{x}_i \) in the source to a transformed point \( \mathbf{x}_i' = \mathcal{T}(\mathbf{x}_i) \) in \( \mathbb{R}^3 \).

The objective is to minimize the alignment error by reducing the difference between transformed points \( \mathbf{x}_i' \) and their corresponding points in \( \mathbf{Y} \). This is achieved by minimizing the following cost function \( E(\mathcal{T}) \):

\begin{equation}
E(\mathcal{T}) = \sum_{i=1}^{N} \| \mathbf{x}_i' - \mathbf{y}_{j(i)} \|^2,
\end{equation}

where \( \| \cdot \| \) denotes the Euclidean distance, and \( \mathbf{y}_{j(i)} \) is the corresponding point in \( \mathbf{Y} \) for each transformed point \( \mathbf{x}_i' \).

\section{Proposed Method} \label{ProposedMethod}
\begin{figure*}[ht]
\centering
\includegraphics[width=18cm]{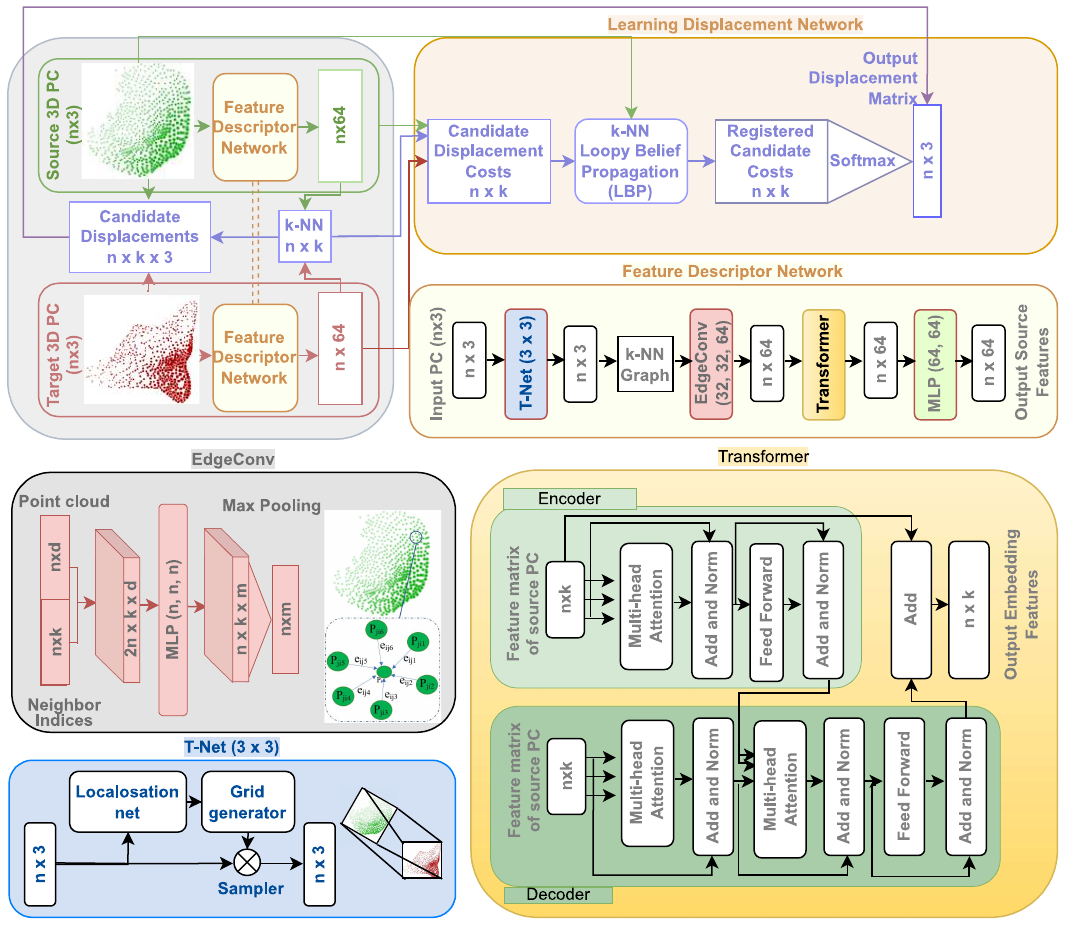}
\caption{The proposed network architecture of DefTransNet}
\label{fig:ProposedMethod}
\end{figure*}

The proposed network in this paper, DefTransNet, is presented in two main subsections: the feature descriptor network and the learning displacement network. DefTransNet is visualized in Figure \ref{fig:ProposedMethod}.

\textit{\textbf{Feature descriptor network.}} The proposed method for feature description involves three key steps. 
The approach begins by using a learnable transformation matrix to align and standardize input point clouds, ensuring feature descriptor invariance to rotations around coordinate axes \cite{jaderberg2015spatial}. A multilayer perceptron (MLP), comprising three 1D convolutional layers (output sizes: 64, 128, 1024), is employed, each followed by instance normalization and ReLU activation. This generates local feature vectors capturing point cloud shape and structure. A max-pooling operation on the 1024 features yields a global feature vector summarizing the entire point cloud. This global vector is fed into another MLP with three linear layers (512, 256, 9 neurons). The final layer outputs a 3×3 matrix representing the transformation matrix, applied to align the input point clouds. \par

The second stage is grounded in the concept of integrating global and local geometric information for aggregating features \cite{wang2019dynamic}. To this end, a local graph is explicitly built, and embeddings for the edges are acquired through learning by using EdgeConv layers \cite{wang2019dynamic}. In this layer, a set of points is taken as input, represented by a matrix $X = \{\mathbf{x}_i \in \mathbb{R}^d : i = 1, 2, ..., n\}$ with dimensions n x d. Here, n signifies the number of points, and d denotes the dimensionality of each point. In addition, a matrix is utilized that encompasses all neighbor indices for each point, reflecting the edges $e$ that connect the points within the graph. The features of the edge $\mathbf{e}_{ij}$ between neighboring points $\mathbf{x}_{i}$ and $\mathbf{x}_{j}$ are calculated as follows:

\begin{equation}
    \mathbf{e}_{ij} = h_{\Theta}\left(\mathbf{x}_i, \mathbf{x}_j\right)=\bar{h}_{\Theta}\left(\mathbf{x}_i, \mathbf{x}_j-\mathbf{x}_i\right),
\end{equation}

where $h_{\Theta}$ acts as a non-linear function with a set of learnable parameters $\Theta=\left(\theta_1, \ldots, \theta_L, \phi_1, \ldots, \phi_L\right)$. Using the coordinates of the patch centers $\mathbf{x}_i$ keeps the global shape structure and $\mathbf{x}_j-\mathbf{x}_i$ captures the local neighborhood information. The operator is defined as
\begin{equation}
e_{i j l}^{\prime}=\operatorname{ReLU}\left(\theta_l \cdot\left(\mathbf{x}_j-\mathbf{x}_i\right)+\phi_l \cdot \mathbf{x}_i\right),
\end{equation}
which can be implemented as a shared MLP, and taking
\begin{equation}
x_{i l}^{\prime}=\max _{j:(i, j) \in e} e_{i j l}^{\prime}.
\end{equation}

The final output is a m-dimensional point cloud with the same number of points as the input point cloud X. 

For the third step of feature generation, the inclusion of a Transformer is proposed to enhance features through embedding learning. Through the use of self-attention and conditional attention, the Transformer can learn novel features by incorporating both source and target features. This capability is particularly advantageous for accurately registering strongly deformed input point clouds, potentially leading to improved accuracy. The model comprises an encoder and a decoder, both constructed from identical building blocks. Each block includes a multi-head self-attention mechanism and a feedforward neural network. The self-attention mechanism is a pivotal feature of the Transformer model, enabling the model to focus on different parts of the input sequence while processing each element. This mechanism computes a weighted sum of the input sequence elements, with weights determined by a similarity score between the input element and all others in the sequence. This empowers the model to assign more significance to crucial elements in the input sequence, a process that can be characterized as

\begin{equation}
    \mathrm{Attention}(Q, K, V) = \mathrm{softmax}(\frac{QK^T}{\sqrt{d_k}})V,
\end{equation}

where $Q$, $K$, and $V$ are the matrices of queries, keys, and values, respectively, and $d_k$ is the dimension of the keys. In each block, the feedforward neural network is utilized to convert the output of the self-attention mechanism into a transformed representation more suitable for the subsequent layer of the encoder or decoder. This transformation is accomplished by employing two linear convolutional layers connected with a ReLU activation function. To optimize the performance of the self-attention mechanism, the Transformer model incorporates various techniques, including multi-head attention. This technique enables the model to concurrently focus on different parts of the input sequence. This can be described as

\begin{equation}
    \mathrm{MultiHead}(Q, K, V) =
    \mathrm{Concat}(\mathrm{head_1}, ..., \mathrm{head_h})W^O,
\end{equation}

$\text{where}~\mathrm{head_i} = \mathrm{Attention}(QW^Q_i, KW^K_i, VW^V_i)$ and $W_i$ are learnable weight matrices, with $i$ between 1 and $h$, where $h$ is the total number of heads. The decoder has one multi-head attention layer more than the encoder, which takes in the result of the encoder and mixes it with the result of the first block in the decoder, before feeding it forward together. In the end, the result of the decoder is added to the initial features that were fed into the encoder, resulting in new features $\Phi_{\mathcal{X,Y}}$. This is done symmetrically for both sets of features $\mathcal{F}_{\mathcal{X}}, \mathcal{F}_{\mathcal{Y}}$

\begin{equation}
\begin{aligned}
& \Phi_{\mathcal{X}}=\mathcal{F}_{\mathcal{X}}+\phi\left(\mathcal{F}_{\mathcal{X}}, \mathcal{F}_{\mathcal{Y}}\right) \\
& \Phi_{\mathcal{Y}}=\mathcal{F}_{\mathcal{Y}}+\phi\left(\mathcal{F}_{\mathcal{Y}}, \mathcal{F}_{\mathcal{X}}\right),
\end{aligned}
\end{equation}
where $\phi$ stands for an asymmetric function given by the Transformer and $X$ and $Y$ for source and target point clouds, respectively \cite{wang2019deep}.

\textit{\textbf{Learning displacement network.}} 
After extracting the relevant features from the graphs, a fully connected multi-layer perceptron (MLP) is employed to generate the output features. Next, the K-Nearest Neighbor (k-NN) algorithm is used to identify the closest \( k \) points in the target point cloud for each point \( \mathbf{x}_{i} \) in the source point cloud \( X = \{\mathbf{x}_i \in \mathbb{R}^3 : i = 1, 2, \dots, n\} \). The squared Euclidean distance is chosen as the metric to find the nearest neighbors. Once the \( k \) nearest neighbors are determined, they serve as candidate points \( \mathbf{c}^{p}_{i} \) for displacements, where \( p \in \{1, 2, \dots, k\} \) for each point \( \mathbf{x}_{i} \) in the source point cloud. These candidates represent possible transformations that align the source point cloud with the target. The displacement cost \( d^{p}_{i} \) for each candidate \( \mathbf{c}^{p}_{i} \) is computed based on their respective feature vectors, reflecting the difference between the feature vector of the source point and the feature vector of the candidate, given by:

\begin{equation}
d^{p}_{i} = \left\| f(\mathbf{x}_{i}) - f(\mathbf{c}^{p}_{i}) \right\|_2^2
\end{equation}

To capture spatial relationships between the points, an additional k-NN graph is constructed for the source point cloud. To address potential errors in registration arising from noisy or missing correspondences, a regularization term \( r_{i j}^{p q} \) is introduced. This term penalizes deviations between the neighboring points \( \mathbf{x}_{i} \) and \( \mathbf{x}_{j} \), and their corresponding candidates \( \mathbf{c}^{p}_{i} \) and \( \mathbf{c}^{q}_{j} \), where \( q \in \{1, 2, \dots, k\} \). This regularization helps enforce consistency in the relative displacements and is defined as:

\begin{equation}
r_{i j}^{p q} = \left\| \left( \mathbf{c}_i^p - \mathbf{x}_{i} \right) - \left( \mathbf{c}_j^q - \mathbf{x}_{j} \right) \right\|_2^2
\end{equation}

The Loopy Belief Propagation (LBP) algorithm is then applied to refine the displacement costs. LBP, a probabilistic inference algorithm, iteratively updates the belief of each point by exchanging information with neighboring points. The outgoing message \( \mathbf{m}_{i \rightarrow j}^t \) from point \( \mathbf{x}_{i} \) to point \( \mathbf{x}_{j} \) in iteration \( t \) is calculated as:

\begin{equation}
\mathbf{m}_{i \rightarrow j}^t = \min_{1, \dots, q, \dots, k} \left( \mathbf{d}_i + \alpha \mathbf{r}_{i j}^q - \mathbf{m}_{j \rightarrow i}^{t-1} + \sum_{(h, i) \in E} \mathbf{m}_{h \rightarrow i}^{t-1} \right)
\end{equation}

Here, \( \alpha \) is a hyperparameter that controls the influence of the regularization term \( \mathbf{r}_{i j}^q = \left( r_{i j}^{1 q}, \dots, r_{i j}^{p q}, \dots, r_{i j}^{k q} \right) \), and the messages are initialized as \( \mathbf{m}_{i \rightarrow j}^0 = 0 \). The term \( \mathbf{m}_{h \rightarrow i}^{t-1} \) represents the messages sent from neighboring points \( \mathbf{x}_h \) in the previous iteration. Finally, the Softmax function is applied to the registered candidate costs, transforming them into weights that reflect the confidence in the registration process. These weighted candidate displacements are then used to compute the final transformation of the source point cloud.

\begin{figure*}[ht]
\centering
\includegraphics[width=14cm]{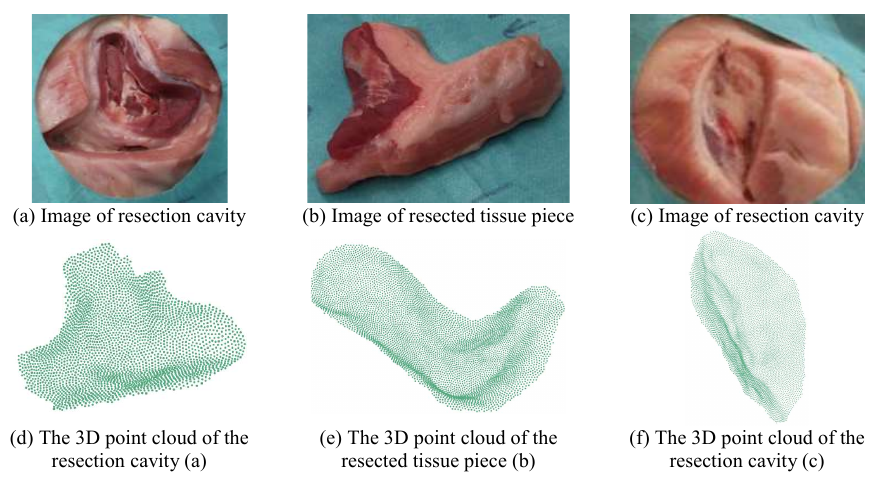}
\caption{Head-mounted display images of resection cavities and cut tissue pieces are shown in the first row and the second row demonstrates the extracted 3D point cloud. The dataset is available upon request to readers.}
\label{fig:DeformedTissueSamples}
\end{figure*}

\section{Evaluation} \label{Evaluation}
Several scenarios are considered to evaluate the robustness and generalization of the proposed DefTransNet. Four datasets are used for this purpose: ModelNet \citep{wu20153d} and SynBench \cite{DataSynBench,monji2024robust} as synthetic datasets, and DeformedTissue \cite{monji2024point} and 4DMatch \cite{li2022lepard} as real-world datasets. In the following subsections, these datasets are introduced, followed by a discussion on the robustness of DefTransNet to various challenges, namely different deformations, noise, outlier levels, and overlap ratios. To this end, the robustness of the proposed method is discussed in comparison to other baselines, namely Robust-DefReg \cite{monji2024robust}, Deep-Geo-Reg \cite{hansen2021deep}, Predator \cite{huang2021predator}, GP-Aligner \cite{wang2022gp}, \cite{croquet2021unsupervised} with regularization, and \cite{croquet2021unsupervised} without regularization.

\subsection{Datasets} 

\textit{\textbf{ModelNet Dataset.}}
One of the datasets used for registration in this study is ModelNet \cite{wu20153d}. The ModelNet10 dataset, a subset of ModelNet40, includes 4,899 pre-aligned shapes across 10 categories, with 3,991 shapes (80\%) designated for training and 908 shapes (20\%) reserved for testing. During the training phase, the models undergo random rotations of up to 45 degrees around the z-axis.

\textit{\textbf{SynBench Dataset.}}
In a previous study, the authors introduced SimTool \cite{monji2023simtool}, a toolbox for simulating soft body deformation and generating deformable point clouds. We utilized SimTool to create SynBench, a benchmark specifically designed for evaluating non-rigid point cloud registration methods. The SynBench dataset, available at \cite{DataSynBench}, consists of five primary sets: "Data," containing 30 primitive objects, and four challenge categories—"Deformation Level," "Incompleteness," "Noise," and "Outlier"—with 5302, 26515, 21213, and 26500 object samples, respectively. Each sample includes a source and target point cloud pair, effectively doubling the total number of files. The increased file count in certain challenges results from applying each challenge across various deformation levels and adjusting effective parameters.\par

The SynBench dataset defines 10 deformation levels, ranging from 0.1 for the lowest level to 1.0 for the highest. As the dataset is synthetically generated, the initial point clouds are noise-free, enabling the controlled addition of synthetic noise. Gaussian noise with zero mean and standard deviations between 0.01 and 0.04 is commonly applied in studies, such as \citep{li2021pointnetlk} and \citep{aoki2019pointnetlk}, to represent small to large noise levels. Accordingly, the noisy dataset is categorized into four groups based on standard deviations: 0.01, 0.02, 0.03, and 0.04 per point set. \par

More information about the SynBench dataset has already been published in our previous paper \cite{monji2024robust}. \par

\textit{\textbf{DeformedTissue Dataset.}}
Tissue deformation is a critical issue in soft-tissue surgery, particularly during tumor resection, as it causes landmark displacement, complicating tissue orientation. The authors conducted an experimental study on 45 pig head cadavers to simulate tissue deformation, approved by the Mannheim Veterinary Office (DE 08 222 1019 21) \cite{monji2024point} \cite{mannle2023artificial}. We used 3D cameras and head-mounted displays to capture tissue shapes before and after controlled deformation induced by heating. The data were processed using software such as Meshroom, MeshLab, and Blender to create and evaluate 2½D meshes. The dataset is available upon request from readers. Some samples of the captured dataset with HoloLens 2 and ArtecEva are shown in Figure \ref{fig:DeformedTissueSamples}. \par

The dataset includes different levels of deformation, noise, and outliers, generated using the same approach as the SynBench dataset. In this paper, the results of the approaches for deformation levels between 0.1 and 0.7 are reported.\par


\begin{table*}[ht]
\footnotesize
\centering
\caption{Mean distance errors across varying deformation levels (0.1 to 0.8) for three datasets: \textit{SynBench (synthetic)}, \textit{ModelNet (synthetic)}, and \textit{DeformedTissue (real world)}. The proposed method \textit{DefTransNet} consistently achieves the lowest mean distance errors compared to other state-of-the-art methods, demonstrating its robustness and accuracy under increasing levels of deformation.
}
\label{tab:DefLevl}
\begin{tabular}{p{0.15cm} p{0.15cm} p{3.05cm} |p{1cm} p{1cm} p{1cm} p{1cm} p{1cm} p{1cm} p{1cm} p{1cm}}
\hline
& & & \multicolumn{8}{|c}{Deformation levels} \\ \cline{4-11}
& &  & \textbf{0.1} & \textbf{0.2} & \textbf{0.3} & \textbf{0.4 }&\textbf{ 0.5} & \textbf{0.6} & \textbf{0.7} & \textbf{0.8} \\ \hline
& & Initial values & 0.03226&	0.07984&	0.12706&	0.20196&	0.28306&	0.33306&	0.36510	&0.42328 \\ 
\multirow{3}{*}{\rotatebox[origin=c]{90}{SynBench}} & \multirow{3}{*}{\rotatebox[origin=c]{90}{Synthetic}} & \textbf{DefTransNet (Ours)} &\textbf{0.00015}&\textbf{0.00037}&	\textbf{0.00062}	&\textbf{0.00122}	&\textbf{0.00646}	&\textbf{0.00763}&	\textbf{0.01972}	&\textbf{0.02110} \\
& & Robust-DefReg(2024) \cite{monji2024robust}& 0.00047 & 0.00116 & 0.00131 &	0.00911&	0.01435&	0.02341	&0.03335&	0.04366\\
& & Deep-Geo-Reg (2021) \cite{hansen2021deep} & 0.00067&0.00141&0.00473&0.01653&0.03194	&0.03975&	0.05281&	0.06122 \\
& & Predator (2021) \cite{huang2021predator}& 0.00051	&0.00358&0.02891&0.04712&0.07123&0.09054&0.13102&0.19821 \\ 
& & GP-Aligner (2022) \cite{wang2022gp} & 0.01816&0.05806&0.07112&0.09106&0.14525&0.17873&0.22912&0.27067\\ 

\hline
\hline
& & Initial values & 0.03485&	0.07055	&0.10776	&0.14317	&0.17576	&0.23051&	0.25216	&0.30652\\ 
\multirow{3}{*}{\rotatebox[origin=c]{90}{ModelNet}} & \multirow{3}{*}{\rotatebox[origin=c]{90}{Synthetic}} & \textbf{DefTransNet (Ours)} & \textbf{0.00078}&	\textbf{0.00138}	&\textbf{0.00409}&	\textbf{0.00488}&	\textbf{0.03196}	&\textbf{0.02791}	&\textbf{0.05641}	&\textbf{0.09889 }\\
& &Robust-DefReg  (2024)\cite{monji2024robust} & \textbf{0.00078}	&\textbf{0.00119}&	0.00492&	0.00619&	\textbf{0.02637}	&0.04118&	0.07644	&0.10707 \\ 
& & Deep-Geo-Reg (2021)\cite{hansen2021deep} & 0.00148&	0.00273&	0.01551&	0.01874	&0.04284	&0.06088&	0.10754	&0.13254 \\ 
& & Predator (2021)\cite{huang2021predator} & 0.00083 & 0.00149 & 0.00591 & 0.01143 & 0.05102&0.08121&0.13124&0.18213 \\ 
& & GP-Aligner (2022)\cite{wang2022gp} &0.02113&	0.03961&	0.05068&	0.07217	&0.11132	&0.15138&	0.17121	&0.21031 \\ 
\hline
\hline
 & & Initial values &0.03551	&0.09090	&0.14890&	0.23681&	0.29240	&0.33700&	0.39258&-
\\ 
\multirow{3}{*}{\rotatebox[origin=c]{90}{DeformedTissue}}& \multirow{3}{*}{\rotatebox[origin=c]{90}{Real-World}} & \textbf{DefTransNet (Ours)}& \textbf{0.00014}&	\textbf{0.00019}&	\textbf{0.00150}&	\textbf{0.00769}	&\textbf{0.01182	}&\textbf{0.01495}&	\textbf{0.01982}&\textbf{-} \\
& &Robust-DefReg (2024)\cite{monji2024robust}&0.00565&	0.01123&	0.02019	&0.07317&	0.08530&	0.09013&	0.09539 & - \\ 
& & Deep-Geo-Reg (2021)\cite{hansen2021deep} &0.00789&	0.01355&	0.02900&	0.09706&	0.09925&	0.10434&	0.11339	&- \\ 
& & Predator (2021)\cite{huang2021predator} & 0.00613&	0.03472&	0.04012&	0.12971&	0.14023&	0.16713&	0.18217&-\\
& & GP-Aligner (2022)\cite{wang2022gp} &0.00925&	0.01932	&0.06120	&0.12057&	0.18014	&0.21632&	0.26423&- \\ 
\hline
\end{tabular}
\end{table*}


\textit{\textbf{4DMatch/4DLoMatch Dataset.}}
4DMatch \citep{li2022lepard} is a benchmark dataset designed for registration and matching tasks, applicable to both rigid and deformable scenes. 4DMatch is a partial point cloud benchmark, while its low-overlap variant is referred to as 4DLoMatch. This dataset is captured using sequences from the \citep{li20214dcomplete} dataset, and it provides ground-truth-dense correspondences. The inclusion of time-varying geometry in both datasets introduces additional challenges for matching and registration tasks.

Each file in the dataset contains the following attributes: the source point cloud \(X\), a deformation array \(D\), the target point cloud \(Y\), the rotation matrix \(R\), the translation vector \(t\), the overlap rate, and the set of corresponding points. For our work, the dataset is modified to generate a new target point cloud \(Y'\) using the formula from \cite{li2022non}, and the original target point cloud is discarded. 

\begin{equation}
Y = t^T + (X+D).R^T
\end{equation}

By applying this transformation only to points with corresponding matches, we ensure that the alignment remains consistent with how the original target was generated, focusing specifically on overlapping points. This approach preserves the integrity of the correspondences, which is essential for registration tasks. Additionally, by ensuring the source and target datasets have an equal number of points, we address a key limitation of our method requiring equal cardinality, making this modification both practical and well-justified.

In the dataset, point clouds with an overlap ratio \(> 0.45\) are labeled as "4DMatch," while those with a lower overlap are labeled as "4DLoMatch." The dataset is split into 47,738 training points, 6,400 validation points, and a test set consisting of 10,327 4DMatch points and 4,590 4DLoMatch points.

\subsection{Robustness to Different Deformation Levels}

Table \ref{tab:DefLevl} presents a comprehensive comparison of the performance of several non-rigid point cloud registration methods, including Robust-DefReg \cite{monji2024robust}, Deep-Geo-Reg \cite{hansen2021deep}, Predator \cite{huang2021predator}, and GP-Aligner \cite{wang2022gp}, across three datasets: SynBench (synthetic), ModelNet (synthetic), and DeformedTissue (real-world). The evaluation metric used is the mean distance error, where smaller values reflect better alignment accuracy. Let \( X = \{x_1, \dots, x_n\} \) and \( Y = \{y_1, \dots, y_n\} \) represent the 3D points in the source and target point clouds, respectively, where \( x_i = (x_i^1, x_i^2, x_i^3) \) and \( y_i = (y_i^1, y_i^2, y_i^3) \). The Euclidean distance between corresponding points \( x_i \) and \( y_i \) is given by:

\[
d(x_i, y_i) = \sqrt{(x_i^1 - y_i^1)^2 + (x_i^2 - y_i^2)^2 + (x_i^3 - y_i^3)^2}.
\]

The \textit{mean distance} \( D_{\text{mean}} \) is then computed as:

\[
D_{\text{mean}} = \frac{1}{n} \sum_{i=1}^{n} d(x_i, y_i).
\]

The deformation levels, ranging from 0.1 to 0.8, progressively introduce higher degrees of non-rigid deformation, allowing for a robust evaluation of each method.

The results demonstrate that the proposed method, DefTransNet, consistently outperforms the state-of-the-art methods across all datasets and deformation levels.

In the SynBench dataset, DefTransNet achieves the lowest mean distance errors across all deformation levels, showcasing its ability to maintain precision even as deformation increases. At a minimal deformation level of 0.1, DefTransNet achieves an error of 0.00015, which is lower than Robust-DefReg (0.00047) and Deep-Geo-Reg (0.00067). As deformation intensifies, DefTransNet remains highly robust, with the error increasing only slightly to 0.02110 at the highest deformation level of 0.8. In contrast, methods such as Predator and GP-Aligner show larger errors, particularly at higher deformation levels, indicating their limitations in handling severe point cloud distortions.

\begin{figure*}[ht]
\centering
\includegraphics[width=18cm]{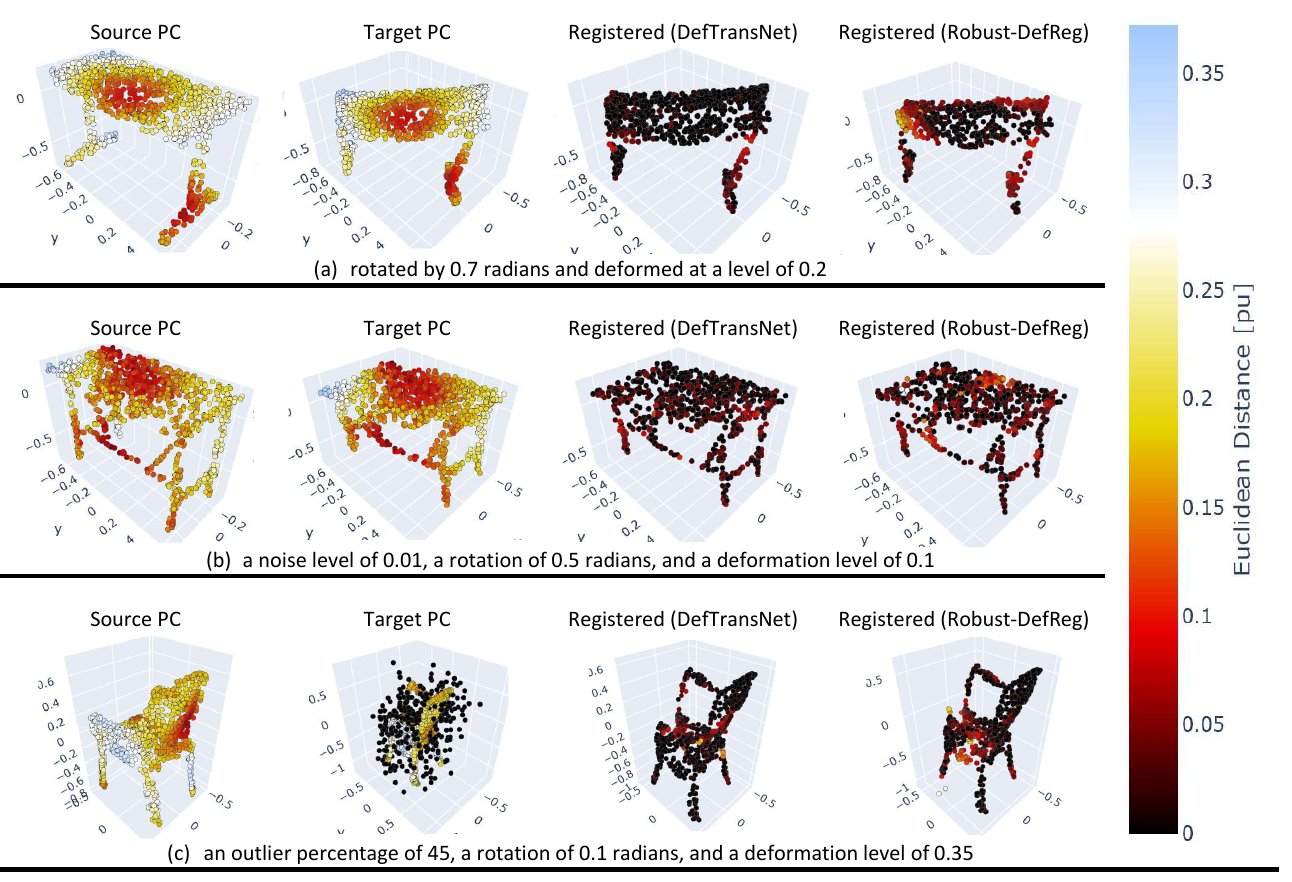}
\caption{Visualization of non-rigid point cloud registration results on ModelNet dataset and under varying conditions: (a) rotation of 0.7 radians and deformation level of 0.2, (b) noise level of 0.01, rotation of 0.5 radians, and deformation level of 0.1, and (c) outlier percentage of 45, rotation of 0.1 radians, and deformation level of 0.35. The results show the source point cloud (Source PC), target point cloud (Target PC), and the registered outputs for DefTransNet and Robust-DefReg. Color coding represents the Euclidean distance error, with lower values shown in darker colors. DefTransNet consistently achieves more accurate alignment, particularly in challenging scenarios with rotations, deformations, noise, and outliers.}
\label{fig:ModelNetVis}
\end{figure*}

A similar performance trend is observed for the ModelNet dataset. DefTransNet again achieves superior accuracy across all deformation levels. For instance, at a moderate deformation level of 0.3, DefTransNet achieves an impressive mean distance error of 0.00078, whereas competing methods such as Deep-Geo-Reg report errors of 0.00148. Even at the extreme deformation level of 0.8, DefTransNet demonstrates remarkable resilience, achieving an error of 0.09988, while the errors of competing methods increase more noticeably. These results highlight DefTransNet’s robustness in accurately registering point clouds, even under challenging synthetic deformations.

The DeformedTissue dataset provides a more challenging evaluation due to the complexities in real-world point clouds. Nevertheless, DefTransNet continues to outperform all other methods, achieving the lowest mean distance errors across all tested deformation levels. At 0.2 deformation, DefTransNet reports an error of 0.00019, whereas Robust-DefReg and Predator yield notably higher errors of 0.01123 and 0.03472, respectively. At the highest tested deformation level of 0.7, DefTransNet maintains its superior performance with an error of 0.01982, significantly outperforming the alternatives. While the mean distance errors for real-world data are generally higher than those observed for synthetic datasets, DefTransNet’s results underscore its ability to handle the inherent complexities of real-world point cloud data effectively.

\begin{figure*}[ht]
\centering
\includegraphics[width=14cm]{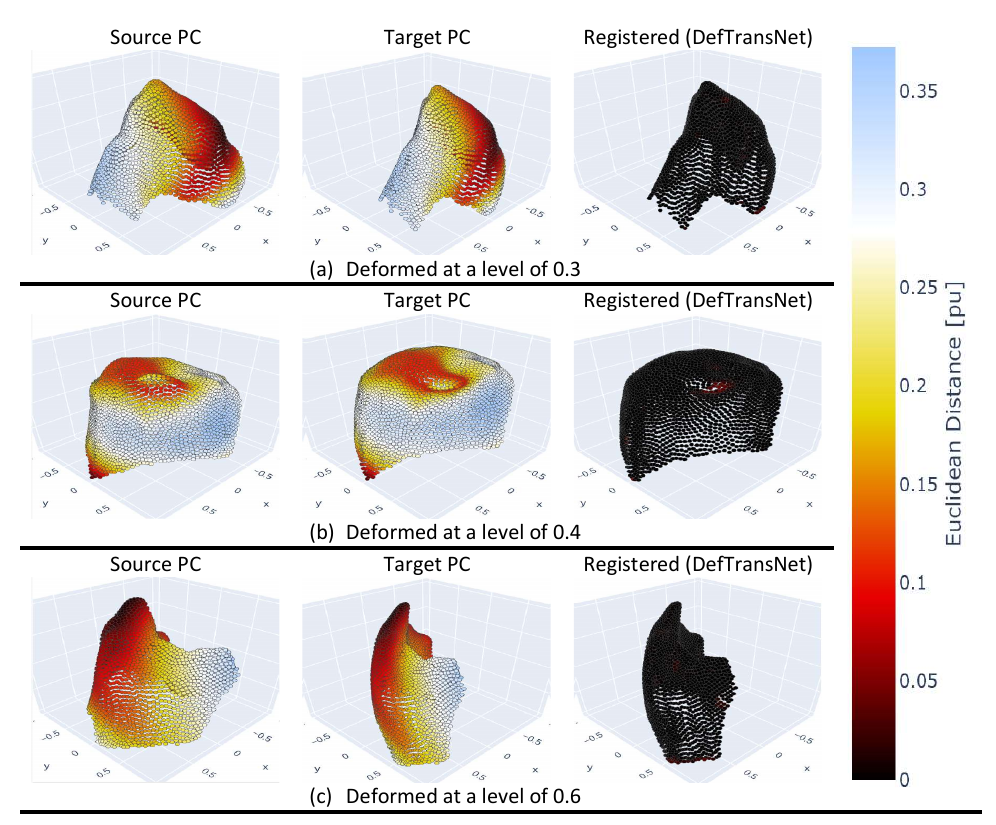}
\caption{Visualization of non-rigid point cloud registration results on the DeformedTissue dataset under increasing deformation levels: (a) 0.3, (b) 0.4, and (c) 0.6. The figure shows the \textit{Source PC} (left), \textit{Target PC} (middle), and the registered outputs using DefTransNet (right). The color bar represents the Euclidean distance error, with darker colors indicating lower registration errors. DefTransNet demonstrates its effectiveness in accurately aligning highly deformed tissue point clouds, even under challenging conditions.}
\label{fig:DeformedTissueVis}
\end{figure*}

The visual results of our approach, in comparison with Robust-DefReg on the ModelNet dataset, are shown in Figure \ref{fig:ModelNetVis}. Furthermore, some visual results of our approach for the DeformedTissue dataset are shown in Figure \ref{fig:DeformedTissueVis}.

\subsection{Robustness to Different Noise and Outlier Degrees}

Table \ref{tab:Noise} evaluates the performance of point cloud registration methods under increasing levels of Gaussian noise (0.01, 0.03, and 0.05) for the SynBench and ModelNet datasets. Noise can distort point positions, posing a challenge to maintaining precise alignments.

\begin{table}[ht]
\footnotesize
\centering
\caption{Mean distance errors under different noise levels (0.01, 0.03, 0.05) for the SynBench and ModelNet datasets. The proposed method DefTransNet consistently outperforms other state-of-the-art methods, demonstrating superior robustness to Gaussian noise.}
\label{tab:Noise}
\begin{tabular}{p{0.1cm} p{0.1cm} p{3cm} |p{1cm} p{1cm} p{0.75cm}}
\hline
& & & \multicolumn{3}{|c}{Noise levels} \\ \cline{4-6}
& & & \textbf{0.01} & \textbf{0.03} & \textbf{0.05} \\ \hline
&&Initial values & 0.27768 & 0.27517 & 0.27824  \\ 
\multirow{3}{*}{\rotatebox[origin=c]{90}{SynBench}} & \multirow{3}{*}{\rotatebox[origin=c]{90}{Synthetic}} & DefTransNet (Ours) & \textbf{0.01544} & \textbf{0.03932} & \textbf{0.06019} \\
& & Robust-DefReg (2024)\cite{monji2024robust} & 0.05393 & 0.06062 & 0.06663 \\ 
& & Deep-Geo-Reg (2021)\cite{hansen2021deep} & 0.07463 & 0.08183 & 0.08562 \\ 
& &Predator (2021)\cite{huang2021predator}& 0.09012&0.09513&0.09921 \\ 
& &GP-Aligner (2022)\cite{wang2022gp} & 0.11387&0.11638&0.12304 \\ \hline \hline 

& &Initial values & 0.22307 & 0.23711 & 0.28056 \\ 
\multirow{3}{*}{\rotatebox[origin=c]{90}{ModelNet}} & \multirow{3}{*}{\rotatebox[origin=c]{90}{Synthetic}} & DefTransNet (Ours) & \textbf{0.01105} & \textbf{0.02065} & \textbf{0.0362}  \\
& & Robust-DefReg (2024)\cite{monji2024robust}  & 0.01075 & 0.02332 & 0.03440  \\ 
& & Deep-Geo-Reg (2021)\cite{hansen2021deep} & 0.02745 & 0.04168 & 0.06289 \\ 
& & Predator (2021)\cite{huang2021predator} &0.05522&	0.05812&	0.07303
 \\ 
& & GP-Aligner (2022)\cite{wang2022gp} & 0.08214	&0.08469	&0.09824\\ \hline 

\end{tabular}
\end{table}

For the SynBench dataset, DefTransNet stands out with the lowest mean distance errors across all noise levels. At a low noise level of 0.01, DefTransNet achieves an error of 0.01544, significantly outperforming methods such as Robust-DefReg (0.05393) and Deep-Geo-Reg (0.07463). Even when the noise increases to 0.05, DefTransNet remains reliable with an error of 0.06019, while other methods, such as Predator and GP-Aligner, show substantial drops in accuracy. These results highlight DefTransNet's ability to suppress the effects of noise and maintain robust performance.

In the ModelNet dataset, DefTransNet continues to perform well, although its results are very close to those of Robust-DefReg. At the smallest noise level (0.01), DefTransNet achieves an error of 0.01105, which is slightly better than Robust-DefReg’s 0.01075. As the noise level increases to 0.05, both methods converge to similar errors, with DefTransNet reporting 0.0362. This indicates that while DefTransNet maintains its robustness, Robust-DefReg also handles noise effectively on this dataset.

Table \ref{tab:Outlier} explores the impact of increasing outlier levels (5\%, 25\%, and 45\%) on registration accuracy. Outliers represent incorrect or unrelated points, making it harder to identify true correspondences between point clouds.

\begin{table}[ht]
\footnotesize
\centering
\caption{Mean distance errors under increasing outlier levels (5, 25, 45\%) for the SynBench and ModelNet datasets. The proposed method DefTransNet achieves the lowest errors, showcasing its robustness to outliers compared to other state-of-the-art methods.}
\label{tab:Outlier}
\begin{tabular}{p{0.1cm} p{0.1cm} p{3cm} | p{1cm} p{1cm} p{0.75cm}}
\hline
& & & \multicolumn{3}{|c}{Outlier levels} \\ \cline{4-6}
& & & \textbf{5} & \textbf{25} & \textbf{45} \\ \hline
&&Initial values & 0.27497 & 0.28185 & 0.27885 \\ 
\multirow{3}{*}{\rotatebox[origin=c]{90}{SynBench}} & \multirow{3}{*}{\rotatebox[origin=c]{90}{Synthetic}} & DefTransNet (Ours) & 	\textbf{0.01388} & \textbf{0.01477} & \textbf{0.01483} \\
& & Robust-DefReg (2024)\cite{monji2024robust}  & 0.05854 & 0.10509 &0.09006 \\ 
& & Deep-Geo-Reg (2021)\cite{hansen2021deep} & 0.07718 & 0.11781 & 0.11361 \\ 
& &Predator (2021)\cite{huang2021predator} & 0.07512&0.11123&0.11591\\ 
& &GP-Aligner (2022)\cite{wang2022gp} & 0.11365&0.13026&0.13642 \\ \hline \hline 

& &Initial values & 0.26762 & 0.28160 & 0.33160 \\ 
\multirow{3}{*}{\rotatebox[origin=c]{90}{ModelNet}} & \multirow{3}{*}{\rotatebox[origin=c]{90}{Synthetic}} & DefTransNet (Ours) & \textbf{0.01303} & \textbf{0.02162} & \textbf{0.03770} \\
& & Robust-DefReg (2024)\cite{monji2024robust}  & 0.04489 & 0.07226 & 0.09570 \\ 
& & Deep-Geo-Reg (2021)\cite{hansen2021deep} & 0.06737 & 0.10861 & 0.13529 \\ 
& & Predator (2021)\cite{huang2021predator}&	0.06312&	0.09832&	0.13101 \\ 
& & GP-Aligner (2022)\cite{wang2022gp}&	0.09132&	0.11036	&0.16069 \\ \hline
\end{tabular}
\end{table}

\begin{figure*}[ht]
\centering
\includegraphics[width=14cm]{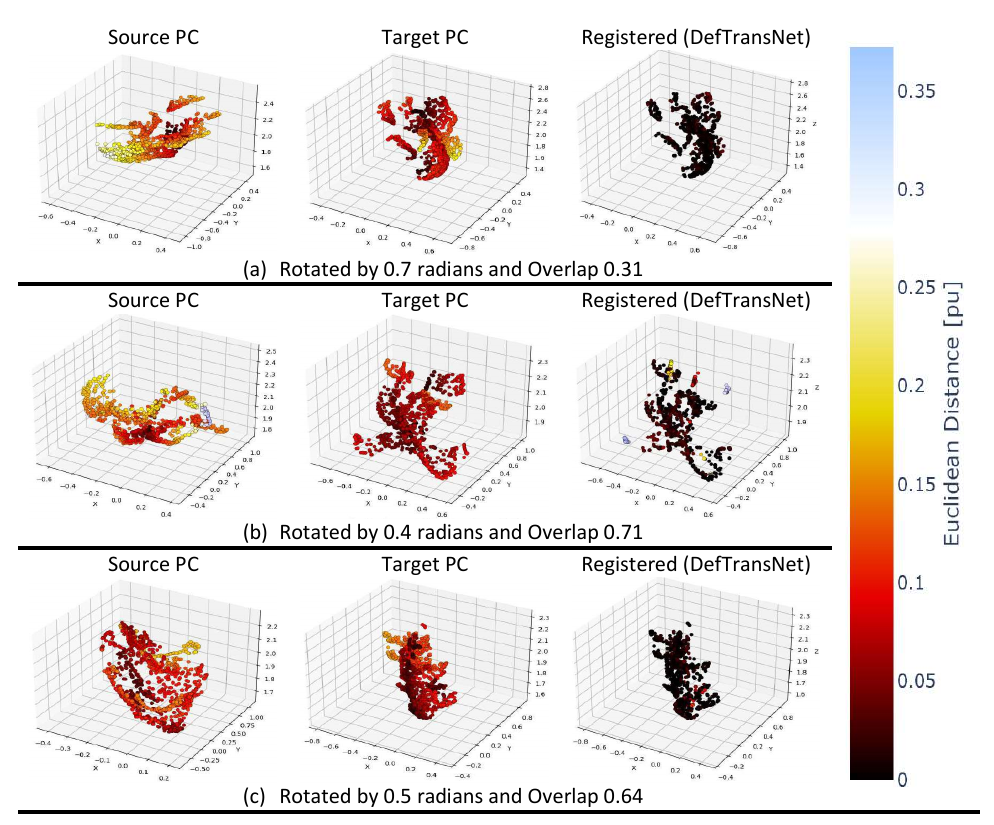}
\caption{Visualization of non-rigid point cloud registration results on the 4DMatch dataset under varying rotation and overlap conditions: (a) rotated by 0.7 radians with 0.31 overlap, (b) rotated by 0.4 radians with 0.71 overlap, and (c) rotated by 0.5 radians with 0.64 overlap. The figure shows the \textit{Source PC} (left), \textit{Target PC} (middle), and the registered outputs using DefTransNet (right). The color bar indicates the Euclidean distance error, with darker colors representing better alignment. DefTransNet successfully handles significant rotations and varying overlaps, producing accurate registration results.}
\label{fig:4DMatchVis}
\end{figure*}

For the SynBench dataset, DefTransNet delivers the best results across all levels of outliers. At a low outlier level of 5\%, DefTransNet achieves a mean distance error of 0.01388, significantly outperforming Robust-DefReg (0.05854) and Deep-Geo-Reg (0.07718). As the outlier percentage increases to 45\%, DefTransNet remains remarkably accurate with an error of 0.01483, while other methods exhibit much larger errors. This shows that DefTransNet effectively distinguishes true points from noisy outliers, maintaining reliable performance even in difficult conditions.

On the ModelNet dataset, a similar trend is observed. At 5\% outliers, DefTransNet achieves an error of 0.01303, far lower than competing methods such as Robust-DefReg (0.04489) and Predator (0.06312). As the level of outliers increases to 45\%, DefTransNet achieves 0.03770, demonstrating its ability to handle high levels of outliers while keeping errors low. The gap between DefTransNet and the competing methods widens as the outlier level increases, emphasizing its strong resilience.

\begin{table*}
\footnotesize
\centering
\caption{Chamfer distance errors under varying overlap ratios (0.1 to 0.9) on the 4DMatch dataset with and without rotational transformations. DefTransNet consistently achieves the lowest errors, demonstrating its robustness to both low overlap and rotational variations compared to baseline methods.}
\label{tab:Overlap}
\begin{tabular}{p{0.25cm} p{0.25cm} p{3cm} |p{1cm} p{1cm} p{1cm} p{1cm} p{1cm} p{1cm} p{1cm} p{1cm} p{1cm}}
\hline
& & & \multicolumn{9}{|c}{Overlap Ratio} \\\cline{4-12}
& & & \
\textbf{0.1} & \textbf{0.2} & \textbf{0.3} & \textbf{0.4 }&\textbf{ 0.5} & \textbf{0.6} & \textbf{0.7} & \textbf{0.8} & \textbf{0.9} \\ \hline
\multirow{4}{*}{\rotatebox[origin=c]{90}{4DMatch}} & \multirow{4}{*}{\rotatebox[origin=c]{90}{With Rotation}} & Initial values & 0.73864	&0.70796&	0.66570	&0.65412	&0.58933&	0.51566	&0.48697	&0.40158&	0.32207 \\
 & & \textbf{DefTransNet (Ours)} & \textbf{0.00520}&	\textbf{0.00522}&\textbf{0.00515}&	\textbf{0.00513}&	\textbf{0.00502}&	\textbf{0.00486}&	\textbf{0.00470}	&\textbf{0.00447}	&\textbf{0.00413} \\
 & &\cite{croquet2021unsupervised} with regularization & 0.15618&	0.17687&	0.17760&	0.18782&	0.17711&	0.15154&	0.17758	&0.15598&	0.13998 \\ 
 & &\cite{croquet2021unsupervised} without regularization & 0.14741&	0.18042&	0.17338&	0.18187&	0.17366&	0.15636	&0.17688&	0.15820	&0.14078  \\ 
\hline
\hline

\multirow{4}{*}{\rotatebox[origin=c]{90}{4DMatch}} & \multirow{4}{*}{\rotatebox[origin=c]{90}{Without Rotation}} & Initial values & 0.87664	&0.70807&	0.67714	&0.69229&	0.57825&	0.48058	&0.43053&	0.32777&	0.28243 \\ 
 & & \textbf{DefTransNet (Ours)} &\textbf{ 0.00530}& 	\textbf{0.00523}	& \textbf{0.00523}	& \textbf{0.00508}& 	\textbf{0.00507}& 	\textbf{0.00478}	& \textbf{0.00455}& 	\textbf{0.00428}& 	\textbf{0.00415} \\
 & &\cite{croquet2021unsupervised} with regularization & 0.17506	&0.21427&	0.16186&	0.22579&	0.17277	&0.12624	&0.16216	&0.17744	&0.13352  \\ 
 & &\cite{croquet2021unsupervised} without regularization & 0.16887&	0.20948&	0.16247&	0.22814&	0.15299&	0.13553&	0.14859&	0.17293&	0.13289 \\ 
\hline
\end{tabular}
\end{table*}

\subsection{Robustness to Different Overlap Ratios}
Table \ref{tab:Overlap} evaluates the performance of DefTransNet compared to other methods under varying overlap ratios (0.1 to 0.9) on the 4DMatch dataset. Two scenarios are considered: with rotation and without rotation. The Chamfer distance error serves as the evaluation metric, where lower values indicate better performance. The Chamfer Distance between two 3D point clouds \( X \) and \( Y \) is defined as:

\[
d_{\text{Chamfer}}(X, Y) = \frac{1}{|X|} \sum_{x \in X} \min_{y \in Y} \|x - y\|_2^2 + \frac{1}{|Y|} \sum_{y \in Y} \min_{x \in X} \|y - x\|_2^2
\]

where \( X \) and \( Y \) represent the two point clouds, \( \|x - y\|_2^2 \) is the squared Euclidean distance between two points \( x \in X \) and \( y \in Y \), and \( \min \) finds the closest point in the opposite point cloud. The Chamfer Distance computes the average squared distance between each point in one point cloud and its nearest neighbor in the other.

Table \ref{tab:Overlap} also includes results for a baseline method referred to as \cite{croquet2021unsupervised} with regularization and \cite{croquet2021unsupervised} without regularization.

The top section of Table \ref{tab:Overlap} presents results when point clouds include rotational transformations. DefTransNet demonstrates superior performance across all overlap ratios. It consistently achieves the lowest mean distance errors, showing its robustness to varying degrees of overlap. For example, at an overlap ratio of 0.1 (low overlap), DefTransNet achieves a mean distance error of 0.00520, significantly outperforming both baselines.
As the overlap increases to 0.9, DefTransNet further refines its alignment accuracy, reporting an error of 0.00413, showcasing its ability to exploit higher overlap effectively. The baseline method \cite{croquet2021unsupervised} with regularization shows moderate performance but lags behind DefTransNet. At an overlap ratio of 0.1, it reports an error of 0.15618, which is notably higher than DefTransNet's result. The performance of \cite{croquet2021unsupervised} without regularization is also less robust, with errors ranging from 0.14741 at 0.1 overlap to 0.14078 at 0.9 overlap. This suggests that regularization could not improve the performance of the baseline, also it remains far behind DefTransNet.

The bottom section of Table \ref{tab:Overlap} evaluates performance when the point clouds have no rotational transformations. DefTransNet again achieves the lowest errors across all overlap ratios. Its performance is slightly better compared to the rotation scenario, indicating that removing rotations simplifies the alignment process. For instance, at an overlap ratio of 0.1, DefTransNet achieves an error of 0.00530, which is only marginally higher than its performance under high overlaps (e.g., 0.00415 at 0.9 overlaps). The trend remains consistent as the overlap increases, with DefTransNet consistently outperforming the other methods. The baseline  \cite{croquet2021unsupervised} with regularization performs better in this case compared to the rotation scenario but still falls short of DefTransNet. At 0.1 overlap, it achieves an error of 0.17506, which is significantly higher than DefTransNet's 0.00530. \cite{croquet2021unsupervised} without regularization continues to show the weakest performance, with errors fluctuating but remaining consistently higher than DefTransNet. Some visual results of our approach are shown in Figure \ref{fig:4DMatchVis}.


\section{Discussion} \label{Discussion}
In this work, we introduced DefTransNet, a Transformer-based framework designed to address the most pressing challenges in non-rigid point cloud registration. The results obtained across multiple datasets—spanning from synthetic datasets like ModelNet and SynBench to more complex real-world datasets such as DeformedTissue and 4DMatch—demonstrate that our approach significantly advances the state of the art. In particular, the comparative evaluations against leading registration methods, including Robust-DefReg \cite{monji2024robust}, Deep-Geo-Reg \cite{hansen2021deep}, Predator \cite{huang2021predator}, and GP-Aligner \cite{wang2022gp}, highlight the exceptional robustness and accuracy of our framework.

A defining characteristic of DefTransNet is its ability to handle large and complex deformations. Previous methods often suffer from feature ambiguity and degraded performance as deformation levels rise. Our results show a clear improvement in alignment accuracy at all tested deformation levels. Even under severe deformations, DefTransNet maintains a stable error profile, outperforming other methods by a considerable margin. This can be attributed to the integration of Transformers, which excels at capturing global context and long-range dependencies, thus producing more discriminative and reliable feature representations. In contrast, earlier approaches relying solely on local geometric cues or traditional iterative optimization struggle to retain robustness as deformations become more pronounced.

Another key advantage introduced by DefTransNet is its improved resilience to noise and outliers. Realistic scenarios, especially in medical applications like soft-tissue surgery, are replete with irregularities due to sensor limitations, occlusions, and partial data. While baseline methods show a significant performance drop under even moderate noise and high outlier levels, our method consistently achieves the lowest mean distance errors. The attention-based feature descriptor in DefTransNet, combined with its learnable transformation matrix, efficiently filters out misleading correspondences. This ensures that noise and outliers exert only a limited impact, enabling the network to lock onto meaningful geometric relationships within the data.

Additionally, our approach demonstrates exceptional performance in scenarios with varying overlap ratios—a pivotal aspect in real-world applications where partial scans, incomplete data, and non-uniform sampling are commonplace. On the challenging 4DMatch benchmark, DefTransNet not only registers point clouds more accurately across a broad range of overlaps but also sustains its advantage under rigid transformations and partial data. This scalability and adaptability make DefTransNet a promising tool for applications like surgical navigation, where the alignment of partially visible and dynamically deforming tissue structures is critical.

The combination of global and local geometric feature extraction through Graph Neural Network layers, followed by the Transformer-based feature descriptor, positions DefTransNet as an end-to-end solution with notable generalization capabilities. The network learns robust embeddings that can align a wide variety of shapes, from synthetic primitives to complex, real-world anatomical structures. Unlike many prior methods that require careful hyperparameter tuning or separate optimization stages to handle complex conditions, DefTransNet adapts seamlessly, simplifying the pipeline from data input to final registration.

It is important to note, however, that this advanced architecture and high-performance outcome come at the cost of increased computational complexity. Transformers, while powerful, can be computationally demanding. Future research could focus on optimizing the architecture and incorporating more efficient attention mechanisms to reduce computational overhead and training times. Furthermore, although we have demonstrated strong generalization on multiple datasets, investigating domain adaptation strategies and transfer learning methods may further improve performance when shifting between drastically different data distributions.

DefTransNet not only advances the current state-of-the-art in non-rigid point cloud registration but also brings forth a robust, scalable, and generalizable approach. By leveraging the strengths of Transformers, global/local feature integration, and end-to-end training, our method consistently outperforms existing techniques in scenarios characterized by severe deformations, high noise, numerous outliers, and low overlap ratios. As a result, DefTransNet opens new avenues for reliable and accurate 3D shape alignment in demanding real-world applications.

\section{Conclusion} \label{Conclusion}
In this paper, we presented DefTransNet, a novel Transformer-based framework for non-rigid point cloud registration. By integrating a robust feature descriptor network and a learning displacement network, DefTransNet addresses critical challenges such as large deformations, noise, outliers, and partial data. Through extensive evaluation of synthetic and real-world datasets, including ModelNet, SynBench, DeformedTissue, and 4DMatch, our method demonstrated superior accuracy and robustness compared to state-of-the-art approaches. The use of Transformers enabled DefTransNet to capture global and local geometric relationships, effectively mitigating feature ambiguity and enhancing registration performance across diverse conditions. With its ability to generalize to unseen data and handle complex scenarios, DefTransNet establishes a significant advancement in non-rigid point cloud registration, paving the way for improved applications in medical imaging, robotics, and beyond. Future work will focus on optimizing computational efficiency and exploring domain adaptation techniques for broader applicability.

\section*{Acknowledgment}\label{ack}
The authors gratefully acknowledge the data storage service SDS@hd supported by the Ministry of Science, Research and the Arts Baden-Württemberg (MWK) and the German Research Foundation (DFG) through grant INST 35/1314-1 FUGG and INST 35/1503-1 FUGG. \par

\bibliographystyle{abbrvnat}
\bibliography{MyRef.bib}
\end{document}